\documentclass[twoside]{article}
\usepackage[accepted]{aistats2015}
\usepackage{amsmath}
\usepackage{amsfonts}
\usepackage{amssymb}
\usepackage{amsthm}
\usepackage{enumerate}
\usepackage{url}
\usepackage{enumitem}
\usepackage{graphicx}
\usepackage[numbers]{natbib} 
\usepackage{epsfig}
\usepackage[lofdepth,lotdepth]{subfig}

\usepackage{xcolor}
\usepackage{hyperref}
\hypersetup{
    colorlinks,
    linkcolor={red!60!black},
    citecolor={blue!75!black},
    urlcolor={blue!80!red}
}

\newcommand{\specialcell}[2][c]{%
  \begin{tabular}[#1]{@{}c@{}}#2\end{tabular}}
\newcommand{\bphi}{{\boldsymbol \phi}}
\newcommand{\bbeta}{{\boldsymbol \beta}}
\newcommand{\gpi}{(\nabla {\phi})^{-1}}
\newcommand{\rb}{\mathcal{R}({\boldsymbol \beta})}

\setlist[itemize]{leftmargin=0.8cm}
\setlist[enumerate]{leftmargin=0.8cm}

\newtheorem{lemma}{Lemma}

\begin{document}


%

%

\twocolumn[

\aistatstitle{Generalized Linear Models for Aggregated Data}

\aistatsauthor{ Avradeep Bhowmik \And Joydeep Ghosh \And Oluwasanmi Koyejo  }

\aistatsaddress{University of Texas at Austin \\ avradeep.1@utexas.edu  \And University of Texas at Austin \\ghosh@ece.utexas.edu  \And Stanford University \\sanmi@stanford.edu } ]

\begin{abstract}
Databases in domains such as healthcare are routinely released to the public in aggregated form. Unfortunately, na\"{i}ve modeling with aggregated data may significantly diminish the accuracy of inferences at the individual level. This paper addresses the scenario where features are provided at the individual level, but the target variables are only available as histogram aggregates or order statistics. We consider a limiting case of generalized linear modeling when the target variables are only known up to permutation, and explore how this relates to permutation testing; a standard technique for assessing statistical dependency. Based on this relationship, we propose a simple algorithm to estimate the model parameters and individual level inferences via alternating imputation and standard generalized linear model fitting. Our results suggest the effectiveness of the proposed approach when, in the original data, permutation testing accurately ascertains the veracity of the linear relationship. The framework is extended to general histogram data with larger bins - with order statistics such as the median as a limiting case. Our experimental results on simulated data and aggregated healthcare data suggest a diminishing returns property with respect to the granularity of the histogram - when a linear relationship holds in the original data, the targets can be predicted accurately given relatively coarse histograms. 
\end{abstract}

\section{Introduction}

Modern life is highly data driven. Datasets with records at the individual level are generated every day in large volumes. This creates an opportunity for researchers and policy-makers to analyze the data and examine individual level inferences. However, in many domains, individual records are difficult to obtain. This particularly true in the healthcare industry where protecting the privacy of patients restricts public access to much of the sensitive data. Therefore, in many cases, multiple Statistical Disclosure Limitation (SDL) techniques  are applied \cite{statisticalcon}. Of these, data aggregation is the most widely used technique \cite{masking}. 

It is common for agencies to report both individual level information for non-sensitive attributes together with the aggregated information in the form of sample statistics. Care must be taken in the analysis of such data, as na\"{i}ve modeling with aggregated data may significantly diminish the accuracy of inferences at the individual level. In particular, inferences drawn from aggregated data may lead to the problem of ecological fallacy \cite{ecologicalfallacy}, hence the resulting conclusions at the group level may be misleading to  researchers and policy makers interested in individual level inferences. An example that has been cited \cite{ludia} is the high correlation between per capita consumption of dietary fat and breast cancer in different countries, which may lead to the incorrect conclusion that dietary fat causes breast cancer \cite{carroll1975}.

Aggregated data in the form of histograms and other sample statistics are becoming more and more common. Further, most of the data that is collected relates to questions for which the respondents have only a few discrete options from which to select their answer. For example, data available from the Generalized Social Survey (GSS) \cite{gss} are often in this form. 
This paper addresses the scenario where features are provided at the individual level, but the target variables are only available as histogram aggregates or order statistics. Despite the prevalence of order-statistic and histogram aggregated data, to the best of our knowledge, this problem has not been addressed in the literature. 

We consider a limiting case of generalized linear modeling when the target variables are only known up to permutation, and explore how this relates to permutation testing~\cite{good2005}; a standard technique for assessing statistical dependency. Based on this relationship, we propose a simple algorithm to estimate the model parameters and individual level inferences via alternating imputation and standard generalized linear model fitting. Our results suggest the effectiveness of the proposed approach when, in the original data, permutation testing accurately ascertains the veracity of the linear relationship. The framework is extended to general histogram data with larger bins - with order statistics such as the median as a limiting case. Our experimental results suggest a diminishing returns property - when a linear relationship holds in the original data, the targets can be predicted accurately given relatively coarse histograms. Our results also suggest caution in in the widespread use of aggregation for ensuring the privacy of sensitive data.


In summary, the main contributions of this manuscript are as follows:
\begin{enumerate}[label = (\roman*)]
\item we propose a framework for estimating the response variables of a generalized linear model given only a histogram aggregate summary by formulating it as an optimization problem that alternates between imputation and generalized linear model fitting.
\item we examine a limiting case of the framework when all the data is known up to permutation. Our examination suggests the effectiveness of the proposed approach when, in the original data, permutation testing accurately ascertains the veracity of the linear relationship.
  \item we examine a second limiting case where only a few order statistics are provided. Our experimental results suggest a diminishing returns property - when a linear relationship holds in the original data, the targets can be predicted accurately given relatively coarse histograms. 
\end{enumerate}
The proposed approach is applied to the analysis of simulated datasets. In addition, we examine the Texas Inpatient Discharge dataset from the Texas Department of State Health Services \cite{txdata} and a subset of the 2008-2010 SynPUF dataset \cite{synpuf}.

\subsection*{Notation}
Matrices are denoted by boldface capital letters, vectors by boldface lower case letters and individual elements of the vector by the same lowercase letter with the boldface removed and the index added as a superscript. $\mathbf{v}^\top$ refers to the transpose of the column vector $\mathbf{v}$. We denote column partitions using semicolons, that is, $\mathbf{M = [X; Y]}$ implies that the columns of the submatrices $\mathbf{X}$ and $\mathbf{Y}$ are, in order, the columns of the full matrix $\mathbf{M}$. We use $\| \cdot \|$ to denote the $L_2$ norm for vectors and Frobenius norm for matrices. The vector $\mathbf{v}$ is said to be in increasing order if $v^{(i)} \leq v^{(j)}$ whenever $i \leq j$, and the set of all such vectors in $\mathbb{R}^n$ is denoted with a subscripted downward pointing arrow as $\mathbb{R}^n_\downarrow$. Two vectors $\mathbf{v}$ and $\mathbf{w}$ are said to be isotonic, $\mathbf{v \sim_\downarrow w}$, if $v^{(i)} \geq v^{(j)}$ if and only if $w^{(i)} \geq w^{(j)}$ for all $i,j$.

\subsection{Preliminaries and Related Work}

Aggregated data is often summarized using a sample statistic, which provides a succinct descriptive summary \citep{wilks1962}. Examples of sample statistics include the average, median and various other quantiles. While the mean is still the most common choice, the best choice for summarizing a sample generally depends on the distribution the sample has been generated from. In many cases, the use of histograms \cite{scott1979} or order statistic summaries is much more ``natural'' e.g. for categorical data, binary data, count valued data, etc. 

The problem of imputing individual level records from the sample mean has been studied in \cite{cudia} and \cite{ludia} among others. In particular, the paper \cite{ludia} attempts to reconstruct the individual level matrix by assuming a low rank structure and compares their framework with other approaches which include an extension of the neighborhood model \cite{neighbourhoodmodel} and a variation of ecological regression \cite{ER} for the task of imputing individual level records of the response variable. However, these approaches exploit the fact that sample mean is a linear function of the sample values. Hence, none of these approaches are extendable to non-linear functions such as order statistics and histogram aggregates. To the best of our knowledge, the special case of individual inferences based on sample statistic aggregate data has not been addressed before. Our work proposes a first solution to address this open problem.

\textbf{Order Statistics:} Given a sample of $n$ real valued datapoints, the $\tau^{th}$ order statistic of the sample is the $\tau^{th}$ smallest value in the sample. For example, the first order statistic is the minimum value of the sample, the $\frac{n}{2}^{th}$ order statistic is the median and the $n^{th}$ order statistic is the maximum value of the sample. 
We specifically design a framework which makes it relatively straightforward to work with order statistics.

\textbf{Histograms :} Given a finite sample of $n$ items from a set $\mathcal{C}$, a histogram is a partition of the set $\mathcal{C}$ into disjoint bins $C_i: \cup_i C_i = \mathcal{C}$ and the respective count or percentage of elements from the sample in each bin. Seen this way, for any sample from $\mathcal{C} \subseteq \mathbb{R}$, a histogram is essentially a set of order statistics for that sample. Histograms can sometimes be specified without their boundary values (eg. "$x < 30$" as opposed to "$0 < x < 30$")- this is equivalent to leaving out the first and the $n^{th}$ order statistic. Further, 
a set of sample statistic summaries are easily converted to (and from) a discrete cumulative distribution by identifying the quantile value as the cumulative histogram boundary, and the quantile identity as the height. This cumulative histogram is easily converted to a standard histogram by differencing of adjacent bins. A similar strategy is also applicable to unbounded domains using abstract \emph{max} and \emph{min} boundaries of $\pm \infty$. Based on this bijection, we we will refer to a histogram as a generalization of the order statistic for the remainder of this manuscript. 

\textbf{Bregman Divergence :} Let $\bphi : \Theta \mapsto \mathbb{R}$ be a strictly convex, closed function on the domain $\Theta \subseteq \mathbb{R}^m$ which is differentiable on int($\Theta$). Then, the Bregman divergence $D_\bphi(\cdot \| \cdot)$ corresponding to the function $\bphi$ is defined as
\begin{equation*}
D_\bphi(\mathbf{y} \| \mathbf{x}) \triangleq \bphi(\mathbf{y}) - \bphi(\mathbf{x}) -  \langle \nabla\bphi(\mathbf{x}), \mathbf{y - x} \rangle
\end{equation*}
From strict convexity, it follows that $D_\bphi(\mathbf{y} \| \mathbf{x}) \geq 0$ and $D_\bphi(\mathbf{y} \| \mathbf{x}) = 0$ if and only if $\mathbf{y = x}$. Bregman divergences are strictly convex in their first argument but not necessarily in their second argument.
In this paper we only consider convex functions of the form $\bphi(\cdot): \mathbf{R}^m \ni \mathbf{x} \mapsto \sum_i \phi(x^{(i)})$ that are sums of identical scalar convex functions applied to each component of the vector $\mathbf{x}$. We refer to this class as \emph{identically separable} (\textbf{IS}). 
Square loss, Kullback-Leibler (KL) divergence and generalized I-Divergence (GI) are members of this family (Table \ref{tb:breg}).

\begin{table}[h]
\begin{center}
\caption{\footnotesize Examples of Bregman Divergences} \label{tb:breg}
\begin{tabular}{| c | c |}
\hline
{$\bphi(\mathbf{x}$)}  &{$D_\bphi(\mathbf{y \| x})$} \\
\hline 
$\frac{1}{2}\|\mathbf{x}\|^2 $  & $\frac{1}{2}\|\mathbf{y - x}\|^2$ \\ \hline
\specialcell{$\sum_i (x^{(i)} \log x^{(i)})$ \\ $\mathbf{x} \in$ Prob. Simplex } & \specialcell{KL($\mathbf{y \| x}$) = \\ $\sum_i \left(y^{(i)} \log(\frac{y^{(i)}}{x^{(i)}})\right)$} \\ \hline
\specialcell{$\sum_i x^{(i)} \log x^{(i)} - x^{(i)}$ \\ $\mathbf{x} \in \mathbb{R}^n_+$}  & \specialcell{GI($\mathbf{y \| x}$) = \\ $\sum_i y^{(i)} \log (\frac{y^{(i)}}{x^{(i)}}) - y^{(i)} + x^{(i)}$}\\ \hline
\end{tabular}
\end{center}
\end{table}

\textbf{Generalized Linear Models: }
While least squares regression is useful for modeling continuous real valued data generated from a Gaussian distribution. This is not always a valid assumption. In many cases, the data of interest may be binary valued or count valued.
A generalized linear model (GLM) \cite{glm72} is a  generalization of linear regression that subsumes various models like Poisson regression, logistic regression, etc. as special cases\footnote{see \cite{nelder1972} or \cite{glm1989} for a detailed discussion on GLMs}. A generalized linear model assumes that the response variables, $y$ are generated from a distribution in the exponential family with the mean parameter related via a link function to a linear function of the predictor $\mathbf{x}$. The model therefore is specified completely by a distribution $P_{\bphi}(\cdot \mid \bbeta)$ from the exponential family, a linear predictor $\eta = \mathbf{x}\bbeta$, and a link function $\gpi(\cdot)$ which connects the expectation parameter of the response variable to the predictor variables as $E(y) = \gpi(\mathbf{x} \bbeta)$. 

As explored in great detail in \cite{banerjee}, Bregman Divergences have a very close relationship with 
generalized linear models. In particular, maximum likelihood parameter estimation for a generalized linear model is equivalent to minimizing a corresponding Bregman divergence. 
For example, maximum likelihood for a Gaussian corresponds to squares loss, for Poisson the corresponding divergence is generalized I-divergence and for Binomial, the corresponding divergence is the KL divergence (see \cite{banerjee} for details).
GLMs have been successfully applied in a wide variety of fields including machine learning , biological surveys \cite{glmbio}, image segmentation and reconstruction \cite{glmimage}, analysis of medical trials \cite{glmtrial}, studying species-environment relationships in ecological sciences \cite{glmeco}, virology \cite{poiss64} and estimating mortality from infectious diseases \cite{glminfect}, among many others, and are widely prized for the interpretability of their results and the extendability of their methods in a plethora of domain specific variations \cite{glmrandom}. They are easy to use and implement and many off-the-shelf software packages are available for most major programming platforms.


\section{Problem Description}
Consider a set of fully observed covariates $\mathbf{X = [x_1; x_2; \cdots ; x_{d-p}]} \in \mathbb{R}^{n \times (d-p)}$, and columns of response variables, $\mathbf{Z = [z_1; z_2; \cdots z_p ]} \in \mathbb{R}^{n \times p}$, which are only known only up to the respective histograms of their values (i.e., up to order statistics).

We assume that each element of $\mathbf{z_i}$ has been generated from covariates $\mathbf{X}$ according to some generalized linear model with parameters $\bbeta_i$. The objective is to estimate the $\bbeta_i$ together with $\mathbf{Z = [z_1; z_2; \cdots z_p ]}$ subject to the given order statistic constraints. Since maximum likelihood estimation in a generalized linear model is equivalent to minimizing a corresponding Bregman divergence, we choose the loss function $\mathcal{L}(\mathbf{Z, \bbeta}) = D_\bphi\left(\mathbf{Z} \| \gpi(\mathbf{X} \bbeta)\right)$ to be minimized over the variables $\mathbf{Z, \bbeta}$ while satisfying order statistics constraints on $\mathbf{Z}$.

Without additional structure, the regression problem for each column can be solved independently, therefore without loss of generality we assume $\mathbf{Z}$ is a single column $\mathbf{z}$. We denote the ${\tau_i}^{th}$ order statistic of $\mathbf{z}$ as ${s_{\tau_i}}$, with $\tau_i \in \{\tau_1, \tau_2, \cdots \tau_h\} \subseteq [n]$, which is the set of $h$ order statistics specified via the histogram. For simplicity, in the following section we consider estimation under a single order statistic which has been computed over the entire column. We extend it subsequently to the more general case of multiple order statistics computed over disjoint partitions. 

Therefore, with Frobenius regularization terms $\mathcal{R}(\mathbf{\bbeta}) = \lambda \|\bbeta\|^2$, the overall problem statement boils down to the following optimization problem:
\begin{equation} \label{eq:orderopt}
\begin{aligned}
&\underset{\mathbf{z}, \bbeta}{\text{min}} & & D_\bphi\left(\mathbf{z} \| \gpi(\mathbf{X \bbeta})\right) + \lambda \|\bbeta\|^2 \\
& \text{s.t. } & & \tau_i^{th}\text{ order statistic of }\mathbf{z_i} = s_{\tau_i}
\end{aligned}
\end{equation}

%

\subsection{Estimation under a Single Order Statistic constraint}

Estimating under order statistics constraints is in general a highly non-trivial problem. It is easy to see that the set of vectors with a given order statistic is not a convex set. Therefore, the above optimization problem looks especially difficult to even represent in a concise manner in terms of $\mathbf{z}$. However, it turns out that with the following reformulation, the analysis of the problem becomes much more manageable.

We rewrite $\mathbf{z = P y}$ where $\mathbf{P} \in \mathbb{P}$ is a permutation matrix and $\mathbf{y}$ is a vector sorted in increasing order. Note the following-
\begin{enumerate}[label = (\roman*)]
\item For a $\mathbf{y} \in \mathbb{R}^n_\downarrow$, if $\mathbf{e^{{\tau_i}}}$ is a row vector with 1 in the ${\tau_i}^{th}$ index and 0 everywhere else, then $\mathbf{e^{{\tau_i}} y}$ represents the ${\tau_i}^{th}$ order statistic of $\mathbf{y}$. Since permutation does not change the value of order statistics, this is also the ${\tau_i}^{th}$ order statistic of $\mathbf{z}$
\item If $\mathbf{\Lambda}$ is the matrix with $\mathbf{\Lambda}_{j,j+1} = -1, \mathbf{\Lambda}_{j,j} = 1$ and $\mathbf{\Lambda}_{j,k} = 0$ for all other $j, k :\ (k - j) \neq 0,\pm 1$, the condition that $\mathbf{y}$ is sorted in increasing order is equivalent to the linear constraint $\mathbf{\Lambda y} \leq 0$.
\end{enumerate}

Putting all this together, the optimization problem (\ref{eq:orderopt}) becomes the following
\begin{equation} \label{eq:yopt}
\begin{aligned}
& \underset{\mathbf{P, y}, \bbeta}{\text{min}}\ \ D_\bphi\left( \mathbf{P y} \| \gpi(\mathbf{X} \bbeta) \right) + \rb \\
& \text{s.t. } \mathbf{e^{{\tau_i}} y} = {s_{\tau_i}},\ \mathbf{\Lambda y} \geq 0,\ \mathbf{P} \in \mathbb{P}
\end{aligned}
\end{equation}

The above optimization problem is jointly convex in $\mathbf{y}$ and $\bbeta$ for a fixed $\mathbf{P}$, but the presence of $\mathbf{P}$ as a variable makes the problem much more complicated. Therefore, we attempt to solve it iteratively for each variable in an alternating minimization framework. The update steps consist of the following for each timestep:
\begin{enumerate}[label = (\roman*)]
\item \resizebox{.42 \textwidth}{!}{$\bbeta_t = \underset{\bbeta}{\text{argmin}}\ D_\bphi\left( \mathbf{P_{t-1} y_{t-1}} \| \gpi(\mathbf{X} \bbeta) \right) + \rb$}

\item $\mathbf{y_t} = \underset{\mathbf{y}}{\text{argmin}} D_\bphi\left( \mathbf{P_{t-1} y} \| \gpi(\mathbf{X} \bbeta_t) \right)$ such that $\mathbf{\Lambda y} \leq 0 $  and $\mathbf{e^{{\tau_i}}y} = {s_{\tau_i}}$ 

\item $\mathbf{P_{t}} = \underset{\mathbf{P} \in \mathbb{P}}{\text{argmin}} \ D_\bphi\left( \mathbf{P y_t} \| \gpi(\mathbf{X} \bbeta_t) \right)$

\end{enumerate}
Step (i) is a standard generalized linear model parameter estimation problem. This problem has been studied in great detail in literature and a variety of off-the-shelf GLM solvers can be used for this. 
We focus instead on steps (ii) and (iii) which are much more interesting. 

For (ii), note that since we assumed that $\bphi$ is identically separable, the same permutation applied to both arguments of the corresponding Bregman divergence $D_\phi(\cdot \| \cdot )$ does not change its value. For any constraint set $\mathcal{C}$, we have $\underset{\mathbf{y} \in \mathcal{C}}{\text{argmin}} \ D_\bphi\left( \mathbf{P y} \| \gpi(\mathbf{X} \bbeta_t) \right) = \underset{\mathbf{y} \in \mathcal{C}}{\text{argmin}} \ D_\bphi\left( \mathbf{y} \| \mathbf{P}^{-1} \gpi(\mathbf{X} \bbeta_t) \right)$ given\footnote{note that for a permutation matrix $\mathbf{P}$, $\mathbf{P^{-1}} = \mathbf{P^\top}$} a fixed $\mathbf{P, X, \bbeta}$. Following this fact, step (ii) is a convex optimization problem in $\mathbf{y}$ and can be solved very easily.

Step (iii) is a non-convex optimization problem in general. However, for an identically separable Bregman divergence it turns out that the solution to this is remarkably simple.

\begin{lemma}\label{lemma:sortP} 
The (set of ) optimal permutation(s) in step (iv) above is given by-\\
\resizebox{.47 \textwidth}{!}{
$
\underset{\mathbf{P} \in \mathbb{P}}{\text{argmin}} \ D_\bphi\left( \mathbf{P y_t} \| \gpi(\mathbf{X} \bbeta_t) \right) = \mathbf{\hat{P}} : \mathbf{\hat{P} y_{t}} \sim_\downarrow \gpi(\mathbf{X} \bbeta_t)
$
}\end{lemma}

In other words, the optimal permutation is the one which makes $\mathbf{y_{i,t}}$ isotonic with $\gpi(\mathbf{X} \bbeta_t)$. Note that the optimal permutation is not unique if $\gpi(\mathbf{X} \bbeta_t)$ is not totally ordered. This is a direct application of the following result which appeared as Lemma 3 in the paper \cite{MR}.

\begin{lemma}\label{lemma:sort}
If $x_1 \geq x_2$ and $y_1 \geq y_2$ and $\bphi(\cdot)$ is identically separable, then
\begin{eqnarray*}
D_\bphi(\bigl[\begin{smallmatrix} x_1 \\ x_2 \end{smallmatrix} \bigr] \| \bigl[\begin{smallmatrix} y_1 \\ y_2 \end{smallmatrix} \bigr]) &\leq & D_\bphi(\bigl[\begin{smallmatrix} x_1 \\ x_2 \end{smallmatrix} \bigr] \| \bigl[\begin{smallmatrix} y_2 \\ y_1 \end{smallmatrix} \bigr])\text{, and } \\[5pt] D_\bphi(\bigl[\begin{smallmatrix} y_1 \\ y_2 \end{smallmatrix} \bigr] \| \bigl[\begin{smallmatrix} x_1 \\ x_2 \end{smallmatrix} \bigr]) &\leq & D_\bphi(\bigl[\begin{smallmatrix} y_2 \\ y_1 \end{smallmatrix} \bigr] \| \bigl[\begin{smallmatrix} x_1 \\ x_2 \end{smallmatrix} \bigr])
\end{eqnarray*}
\end{lemma} 

\subsubsection{Solution in terms of $\mathbf{z}$}

Lemmata \ref{lemma:sortP} and \ref{lemma:sort} suggest that we can optimize jointly over $\mathbf{P}$ and $\mathbf{y}$ instead of separately, since for any $\mathbf{y}$ we already know the optimal $\mathbf{P}$. Combining the optimization steps (ii) and (iii) in terms of $\mathbf{P}$ and $\mathbf{y}$, our update step for $\mathbf{z}$ in the original optimization problem is the following 
\begin{eqnarray}\label{eq:updatez}
\mathbf{\hat{z}_{t}} &=& \underset{\mathbf{z}}{\text{argmin}} \ D_\bphi\left( \mathbf{z} \| \gpi(\mathbf{X} \bbeta_t) \right) \\ &\text{s.t.}& \tau_i^{th}\text{ order statistic of }\mathbf{z} = s_{\tau_i} \nonumber
\end{eqnarray}

It is not immediately obvious how to approach the solution to this since the constraint set for $\mathbf{z}$ is not convex. However, note that as a result of Lemma \ref{lemma:sort} it is clear that given a fixed $\mathbf{X}$ and $\mathbf{\bbeta_t}$ if $\mathbf{\hat{z}_{t}}$ is a solution to the subproblem (\ref{eq:updatez}), we must have $\mathbf{\hat{z}_{t}} \sim_\downarrow \gpi(\mathbf{X} \bbeta_t)$. 

Therefore, instead of searching over the set of all vectors in $\mathbb{R}^n$, it is sufficient to search only in the subset of vectors that are isotonic with $\gpi(\mathbf{X} \bbeta_t)$. It turns out that not only is this set convex given a fixed $\mathbf{X}, \bbeta_t$, the solution for $\mathbf{z}_t$ is readily available in closed form.

Let $\mathbf{\Gamma_{t} =} \gpi(\mathbf{X} \bbeta_t)$. Since the Bregman Divergence is IS, without loss of generality we can assume that $\mathbf{\Gamma_{t}}$ is in increasing order, therefore the constraint set for $\mathbf{z}$ becomes $\mathbf{z} \in \mathbb{R}^n_\downarrow$ and order statistics constraints for $\mathbf{z}$ becomes the linear constraint $\mathbf{e^{{\tau_i}}z} = s_{\tau_i}$. 

Therefore, the optimization problem (\ref{eq:updatez}) over $\mathbf{z}$ is equivalent, up to a simple re-permutation step, to the following
\begin{equation}
\begin{aligned}\label{eq:zfinal}
\underset{\mathbf{z}}{\text{min}}  &\ D_\phi(\mathbf{z \| \Gamma_{t}} ) \\
\text{s.t. } &\mathbf{z} \in \mathbb{R}^n_{\downarrow},\  \mathbf{e^{{\tau_i}}z} = {s_{\tau_i}} \\
\end{aligned}
\end{equation}

\begin{lemma}\label{lemma:zfinal} Let $\mathbf{\hat{z}}$ be the solution to the optimization problem (\ref{eq:zfinal}). Then, $\mathbf{\hat{z}}$ is given by-
\begin{equation}\label{eq:zhat}
 \hat{z}_{t}^{(j)} =
  \begin{cases}
   s_{\tau_i}      & j = \tau_i \\
   \max(\Gamma_{t}^{(j)}, s_{\tau_i})  & j > \tau_i  \\
   \min(\Gamma_{t}^{(j)}, s_{\tau_i})  & j < \tau_i
  \end{cases}
\end{equation}
\end{lemma}
\textbf{Sketch of Proof} In the space of all $\mathbf{z}$ ordered in increasing order, the $\tau_i^{th}$ order statistic constraint simply becomes $\hat{z}_{t}^{(j)} < s_{\tau_i}$ for $j < \tau_i$ and vice versa for $j > \tau_i$. Suppose we were to optimize over all space instead of $\mathbb{R}^n_\downarrow$- because the Bregman divergence is identically separable, the optimization problem separates out over different coordinates $j$ as $\hat{z}_{t}^{(j)} = \text{arg}\min_{z} D_\phi(z \| \Gamma_{t}^{(j)})$ such that $z < (>) s_{\tau_i}$ for $j < (>) \tau_i$. This is a unidimensional convex optimization problem the solution to which is given by equation (\ref{eq:zhat}) above.

Finally we note that $\mathbf{\hat{z}_{t}}$, automatically lies in $\mathbb{R}^n_\downarrow$ since $\mathbf{\Gamma_{t}} \in \mathbb{R}^n_\downarrow$, and hence, is also the solution to the optimisation problem (\ref{eq:zfinal}). $\square$


Now, note that since we are performing iterative minimization, the cost function is non-increasing at every step. As the cost function is bounded below by 0, the algorithm converges to a stationary point. We now extend the framework to include histogram constraints and blockwise partitioning.

\subsection{Histogram Constraints} 
In case there are multiple order statistics constraints (histogram), the solution can be obtained by repeated application of equation (\ref{eq:zhat}). 

Suppose for the column $\mathbf{z}$ we have constraints as ${\tau_i}^{th}$ order statistic of $\mathbf{z} = s_{\tau_i}$  for $\tau_i \in \{\tau_1, \tau_2, \cdots \tau_h\} \subseteq \{1, 2, \cdots n\}$, the solution is given by the following-
\begin{enumerate}
\item For all $j < \tau_1$, $\hat{z}^{(j)} = \min(\Gamma_i^{(j)}, s_{\tau_1})$; similarly, for all $j > \tau_h$, $\hat{z}^{(j)} = \max(\Gamma_i^{(j)}, s_{\tau_h})$
\item For all $1 \leq k < h$, and $j\ : \tau_k \leq j \leq \tau_{k + 1}$,
\begin{equation*}
 \hat{z}^{(j)} =
  \begin{cases}
   s_{\tau_k}      & j = \tau_k \\
   s_{\tau_{k+1}}      & j = \tau_{k+1} \\
   \min\left(s_{\tau_{k+1}}, \max(\Gamma_i^{(j)}, s_{\tau_k})\right)  & \tau_k \leq j \leq \tau_{k + 1}
  \end{cases}
\end{equation*}
\end{enumerate}

The proof for this follows in an identical manner to the proof for the non-partitioned case earlier. As above, the updated $\mathbf{z_t}$ can be obtained by re-permuting $\mathbf{\hat{z}}$ to preserve isotonicity with $\gpi(\mathbf{X}\bbeta)$. For a fully observed histogram, the update for $\mathbf{z}$ only involves a permutation at each step.

\subsection{Blockwise Order Statistic Constraints} 
In the setup where the order statistics (or histograms) are computed over blockwise partitions of the sample, the permutation matrix is a blockwise permutation matrix and the isotonicity constraint is a blockwise isotonicity costraint. 

Since the Bregman Divergence is identically separable, the update for $\mathbf{z}$ separates out into independent updates for every block which can be done in a manner identical to that given by Lemma \ref{lemma:zfinal}. The update step for $\bbeta$ remains unchanged.

\begin{figure*}[!htb]
\centering
\subfloat[subfig1][Poisson Fit Error]{
\includegraphics[width=0.33\textwidth]{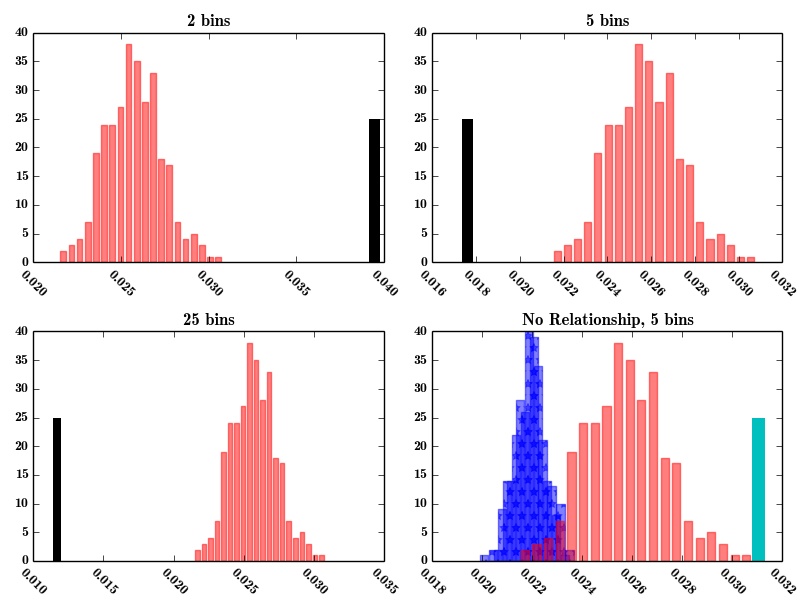}
\label{fig:PoiPtrain}}
\subfloat[subfig2][Gaussian Fit Error]{
\includegraphics[width=0.33\textwidth]{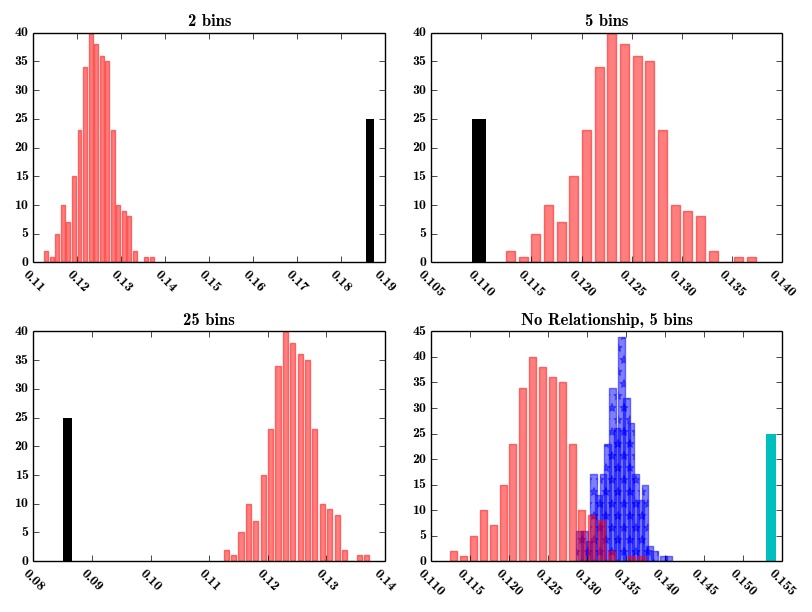}
\label{fig:GauPtest}}
\subfloat[subfig3][Binomial Fit Error]{
\includegraphics[width=0.33\textwidth]{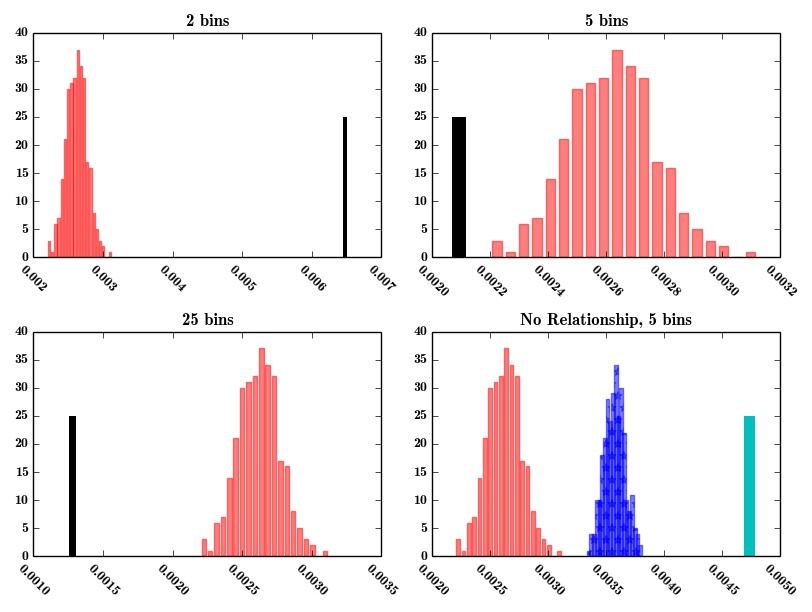}
\label{fig:BinPtest}}
\caption{\footnotesize Permutation tests under Poisson, Gaussian and Binomial Estimation for 2, 5, 25 bins (top left, top right, bottom left) and "No Relationship" (bottom right)}
\label{fig:Ptest}
\vspace*{-0.5cm}
\end{figure*}

\begin{figure*}[!htb]
\centering
\subfloat[subfig1][Poisson Training Error]{
\includegraphics[width=0.33\textwidth , height = 4.25cm]{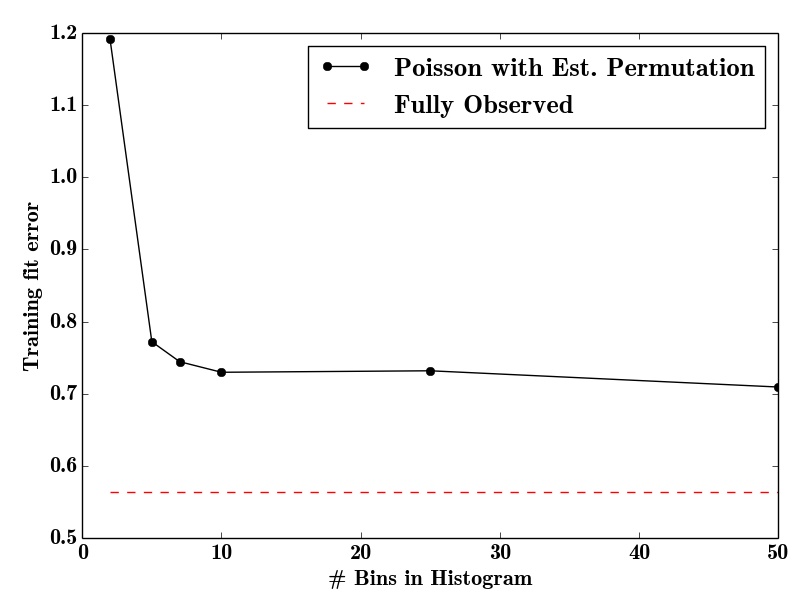}
\label{fig:Poitrain}}
\subfloat[subfig2][Gaussian Training Error]{
\includegraphics[width=0.33\textwidth]{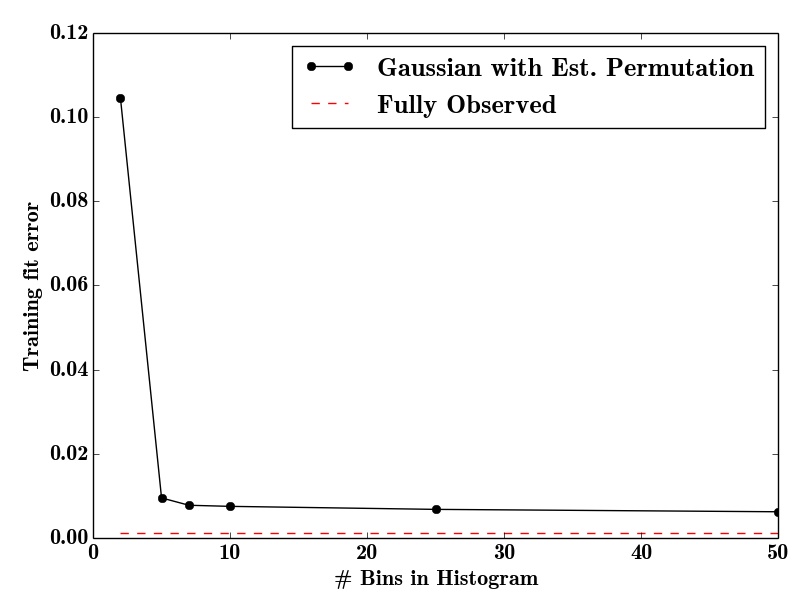}
\label{fig:Gautrain}}
\subfloat[subfig3][Binomial Training Error]{
\includegraphics[width=0.33\textwidth , height = 4.25cm]{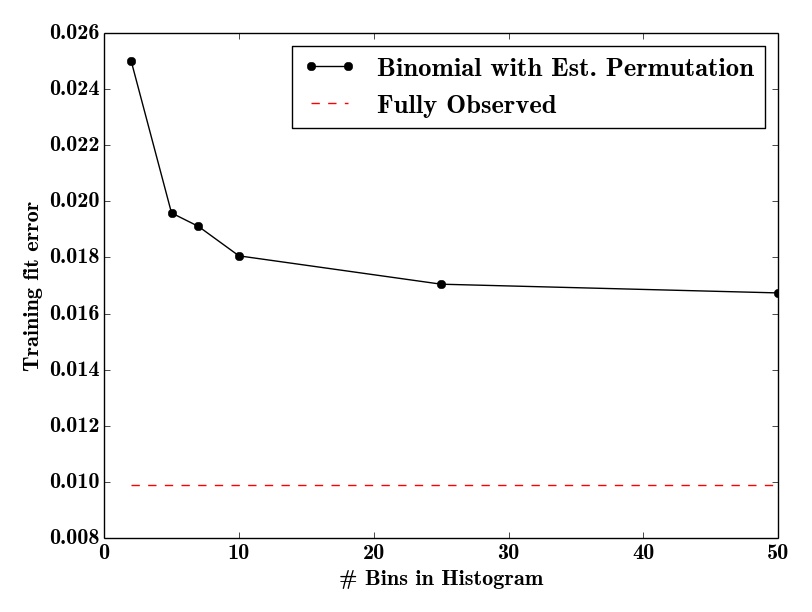}
\label{fig:Bintrain}}
\caption{Training Error under Poisson, Gaussian and Binomial Estimation}
\label{fig:simtrain}
\vspace*{-0.5cm}
\end{figure*}

\begin{figure*}[!htb]
\centering
\subfloat[subfig1][Poisson Test Error]{
\includegraphics[width=0.33\textwidth, height = 4.25cm]{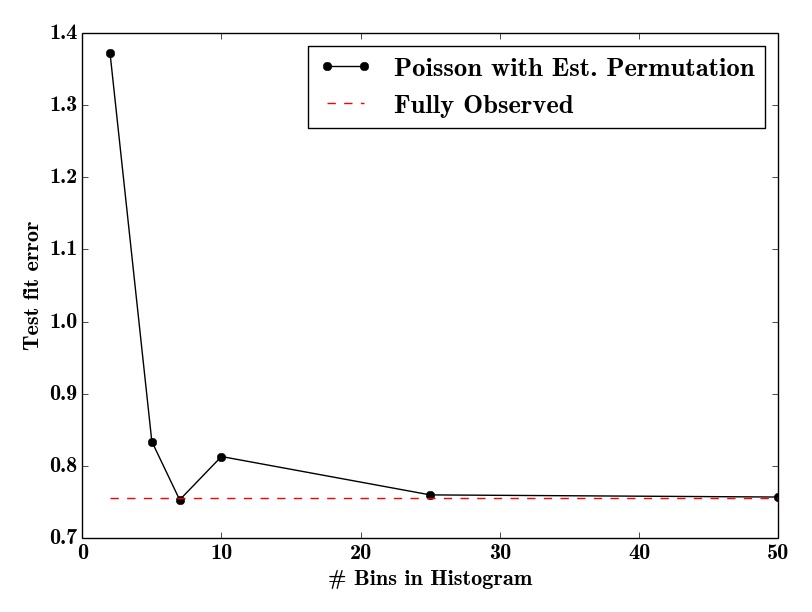}
\label{fig:Poitest}}
\subfloat[subfig2][Gaussian Test Error]{
\includegraphics[width=0.33\textwidth]{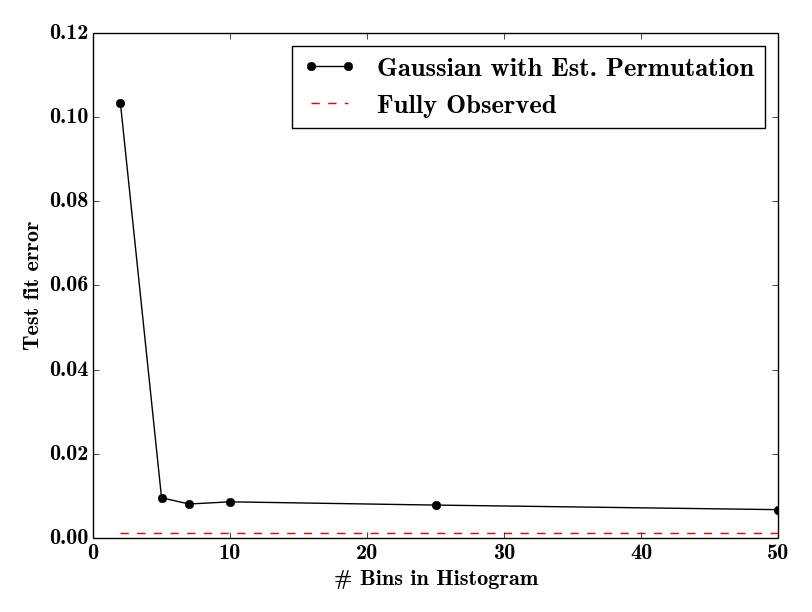}
\label{fig:Gautest}}
\subfloat[subfig3][Binomial Test Error]{
\includegraphics[width=0.33\textwidth, height = 4.25cm]{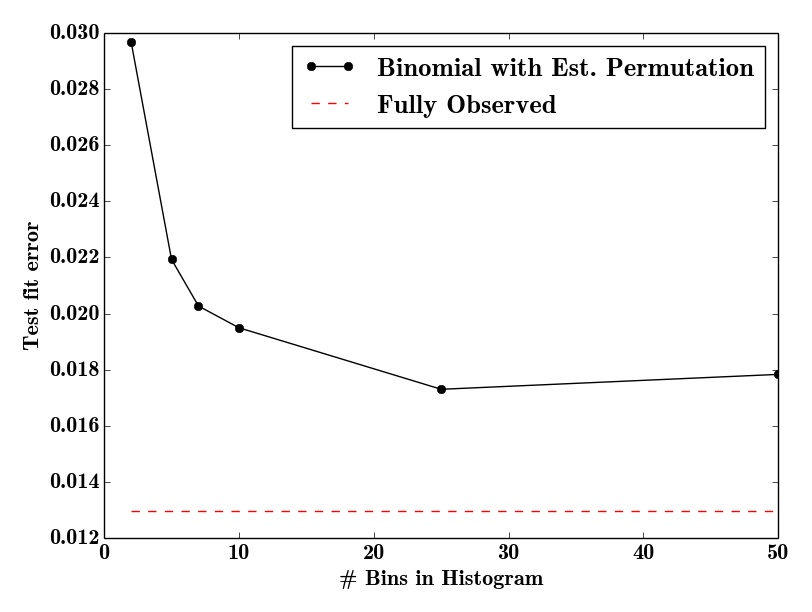}
\label{fig:Bintest}}
\caption{Test Set Error under Poisson, Gaussian and Binomial Estimation}
\label{fig:simtest}
\vspace*{-0.5cm}
\end{figure*}

\begin{figure*}[!htb]
\centering
\subfloat[subfig1][Training Set Error]{
\includegraphics[width=\columnwidth]{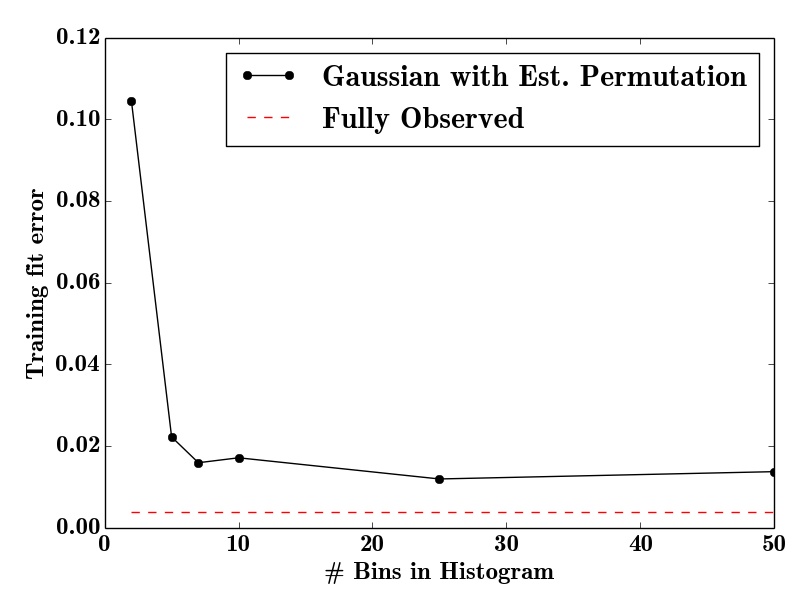}
\label{fig:syntrain}}
\subfloat[subfig2][Test Set Error]{
\includegraphics[width=\columnwidth]{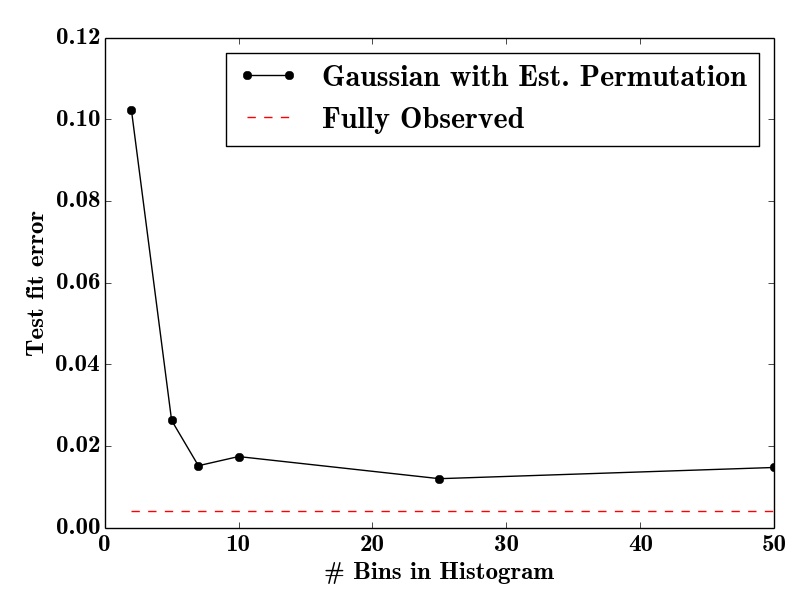}
\label{fig:syntest}}
\caption{Performance on SynPUF dataset}
\label{fig:synpuf}
\vspace*{-0.5cm}
\end{figure*}

\begin{figure*}[!htb]
\centering
\subfloat[subfig1][Training Set Error]{
\includegraphics[width=\columnwidth]{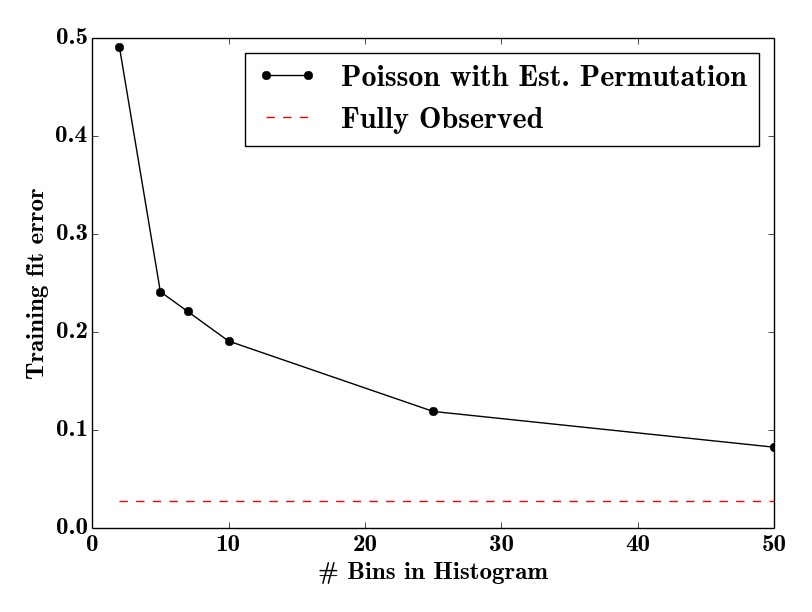}
\label{fig:txdtrain}}
\subfloat[subfig2][Test Set Error]{
\includegraphics[width=\columnwidth]{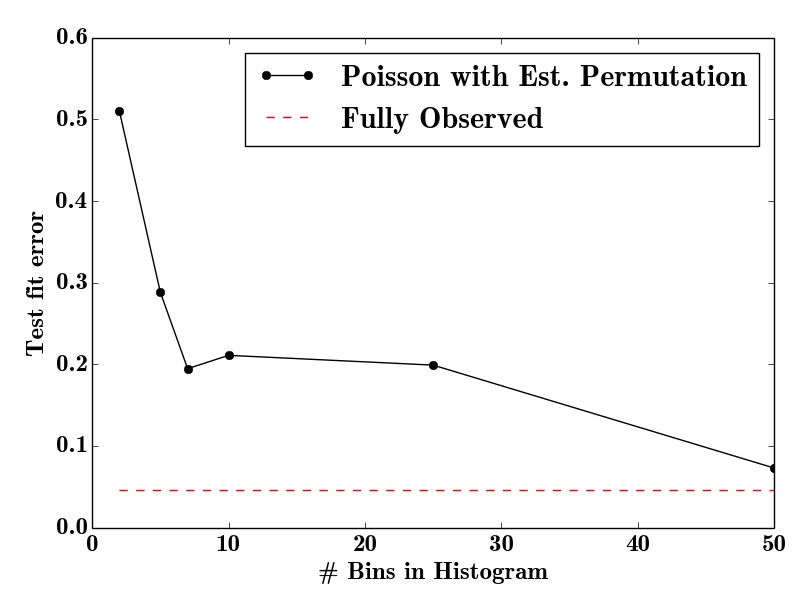}
\label{fig:txdtest}}
\caption{Performance on Texas Inpatient Discharge dataset}
\label{fig:txd}
\vspace*{-0.5cm}
\end{figure*}

\begin{figure*}[!htb]
\centering
\subfloat[subfig1][Recovered Histogram of DE-SynPUF Data]{
\includegraphics[width=\columnwidth , height = 3.5cm]{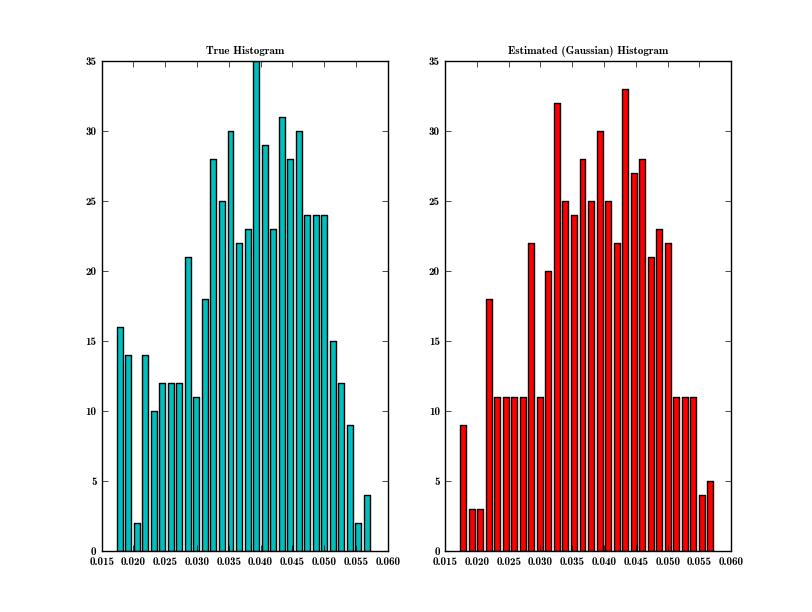}
\label{fig:synpufhist}}
\subfloat[subfig2][Recovered Histogram of Texas Discharged Data]{
\includegraphics[width=\columnwidth , height = 3.3cm]{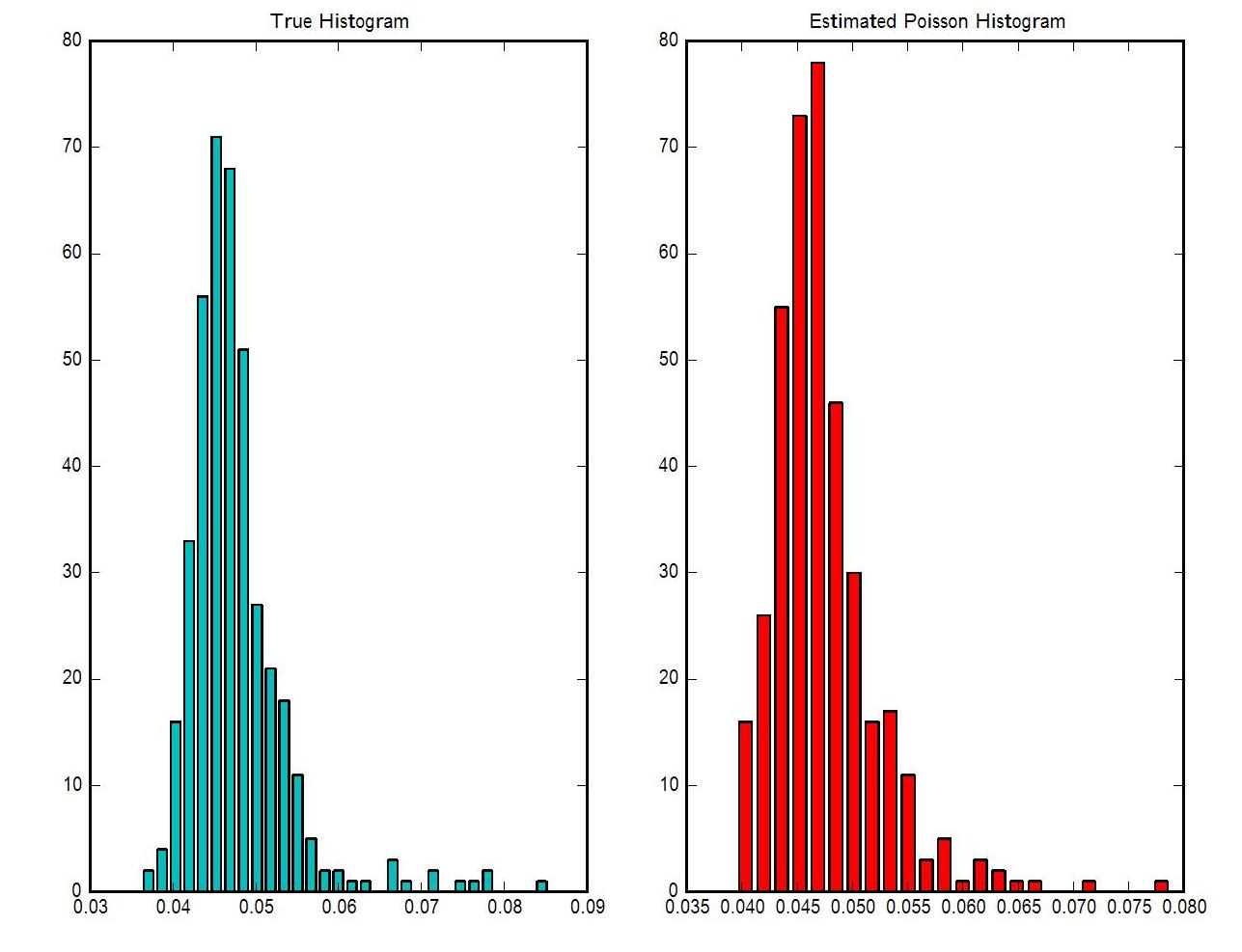}
\label{fig:txdhist}}
\caption{Recovered Histograms of both datasets (true histograms on the left)}
\label{fig:hist}
\vspace*{-0.5cm}
\end{figure*}

\section{Experiments}

We provide experimental results using both simulated data and real data. Error for each generalized linear model is defined as the corresponding Bregman divergence (square loss for Gaussian, generalized I-divergence for Poisson, etc. see \cite{banerjee}) between the true and recovered targets. The average errors for each model is shown separately.

\subsection{Simulated Data}
We randomly generate different sets of real valued predictor variables and parameters, and use the corresponding exponential family to generate their respective response variables. We compute histograms for the response variables thus generated with varying number of bins and test our algorithm for each case. We perform the experiments for three different models - Gaussian, Poisson and Binomial.

We perform a basic permutation test\footnote{Refer to \cite{good2005} for more details on permutation tests} to show how our algorithm performs with respect to the fit by a generalized linear model which knows the values of the target variables but permutes the target variables randomly for estimation. We perform the randomized permutations multiple times and plot a histogram of the fitting errors thus obtained and see how the results from our algorithm compares to the histogram (Figure \ref{fig:Ptest}). The black bar is the error obtained by our framework, the red histogram is the histogram of errors obtained by fitting after randomly permuting the targets. The blue histogram is the histogram of errors obtained by fitting a model where there is no relationship between the target variable and the covariate, the cyan bar is the result of our framework applied to this data with a histogram of 5 bins (histograms of other granularities perform similarly). Our test successfully rejects the null hypothesis of ``no relationship” when the the black bar is to the left of the red histogram. Figure \ref{fig:Ptest} shows that as histogram becomes finer (i.e number of bins increase) error is lower i.e. black bar shifts towards left.

We plot the average fitting and predictive performance of our algorithm with increasing number of bins over five fold cross validation. We compare our results with the results obtained with the best possible GLM estimator which observes the full dataset (Figures \ref{fig:simtrain} and \ref{fig:simtest})\footnote{training/test error in figures \ref{fig:Gautrain} and \ref{fig:Gautest} for the Gaussian estimator for the fully observed case is $\approx 0$}. It can be seen in each case that as the histogram of targets becomes finer (i.e., more bins) the error decreases but with a diminishing returns property with respect to the coarseness of the histogram.


\subsection{DE-SynPUF dataset}

The CMS Beneficiary Summary DE-SynPUF dataset is a public use dataset created by the Centers for Medicare and Medicaid Services by applying different statistical disclosure limitation techniques to real beneficiary claims data in a way so as to very closely resemble real Medicare data. It is often used for testing different data mining or statistical inferential methods before getting access to real Medicare data. We use a subset of the DE-SynPUF dataset for a single state from the year 2008. With some trimming of datapoints (eg, we do not take into account deceased beneficiaries) we model outpatient institutional annual primary payer reimbursement amount (\emph{PPPYMT-OP}) with a number of available predictor variables including age, race, sex, duration of coverage, presence/absence of a variety of chronic conditions, etc.

We perform a log transform and compute histograms of varying granularity on the target variables. We use a Gaussian model for our estimation and evaluate the average performance of our algorithm over five fold cross validation in fitting both the training and test data sample points, comparing with the best possible Gaussian estimator which performs the estimation by observing the full dataset (Figures \ref{fig:synpuf}). As seen in the plot, the performance of our framework improves as the histogram of targets becomes finer in granularity and approaches the performance of the best Gaussian estimator. We also compare the histogram of target variables as recovered by our framework with the true histogram (Figure \ref{fig:synpufhist}).

\subsection{Texas Inpatient Discharge dataset}
We then test our algorithm on the Texas Inpatient Discharge dataset from the Texas Department of State Health Services \cite{txdata} used in \cite{ludia}. As with the simulated data, we use histograms of varying granularity on the respective response variables and evaluate the average performance in fitting both the training and test data sample points over five fold cross validation. We use hospital billing records from the fourth quarter of 2006 in the Texas Inpatient Discharge dataset and regress it on the available individual level predictor variables including binary variables race and sex, categorical variables county and zipcode, and real valued variables like length of stay. 

Following \cite{ludia}, we perform a log transform on the hospital charges and length of stay before applying a Poisson regression model. We compare the performance of our algorithm over five-fold cross-validation with the best possible Poisson estimator which estimates in a fully observed scenario with an uncensored dataset (Figure \ref{fig:txd}). The plot shows that the performance of our framework improves with increasingly finer granularity of histograms and approaches the performance of the best Poisson estimator. Finally, we compare the histogram recovered by our framework with the true histogram for the dataset (Figure \ref{fig:txdhist}).

\section{Conclusion and Future Work}
This paper addresses the scenario where features are provided at the individual level, but the target variables are only available as histogram aggregates or order statistics. We proposed a simple algorithm to estimate the model parameters and individual level inferences via alternating imputation and standard generalized linear model fitting. We considered two limiting cases. In the first, the target variables are only known up to permutation. Our results suggest the effectiveness of the proposed approach when, in the original data, permutation testing accurately ascertains the veracity of the linear relationship. The framework was then extended to general histogram data with larger bins - with order statistics such as the median as a second limiting case. Experimental results on simulated data and real healthcare data show the effectiveness of the proposed approach which may have implications on using aggregation as a means of preserving privacy. For future work, we plan a more detailed analysis to better understand the properties and limits of the framework given binned histogram data. We also plan to extend the approach to non-linear modeling.

\subsubsection*{Acknowledgements}
Authors acknowledge support from NSF grant IIS 1421729.

\bibliographystyle{abbrv}
\bibliography{samplestats}

\end{document}